%% file: paper.tex
\documentclass[10pt,twocolumn,letterpaper]{article}

\usepackage{wacv}
\usepackage{times}
\usepackage{epsfig}
\usepackage{graphicx}
\usepackage{amsmath}
\usepackage{amssymb}
\usepackage{booktabs}
\usepackage{subfig}
\wacvfinalcopy % *** Uncomment this line for the final submission

\def\wacvPaperID{931} % *** Enter the wacv Paper ID here

\ifwacvfinal\fi
\begin{document}

%%%%%%%%% TITLE
\title{Cloud Removal in Satellite Images Using Spatiotemporal Generative Networks}

% Authors at the same institution
%\author{First Author \hspace{2cm} Second Author \\
% Authors at different institutions
\author{Vishnu Sarukkai \\
Computer Science Department\\
Stanford University\\
\and
Anirudh Jain \\
Computer Science Department\\
Stanford University\\
\and
Burak Uzkent \\
Computer Science Department\\
Stanford University\\
\and
Stefano Ermon \\
Computer Science Department\\
Stanford University\\
}

\maketitle
\ifwacvfinal\thispagestyle{empty}\fi

%%%%%%%%% ABSTRACT
\begin{abstract}
   Satellite images hold great promise for continuous environmental monitoring and earth observation. Occlusions cast by clouds, however, can severely limit coverage, making ground information extraction more difficult. Existing pipelines typically perform cloud removal with simple temporal composites and hand-crafted filters. In contrast, we cast the problem of cloud removal as a conditional image synthesis challenge, and we propose a trainable spatiotemporal generator network (\emph{STGAN}) to remove clouds. We train our model on a new large-scale spatiotemporal dataset that we construct, containing 97640 image pairs covering all continents. We demonstrate experimentally that the proposed \emph{STGAN} model outperforms standard models and can generate realistic cloud-free images with high PSNR and SSIM values across a variety of atmospheric conditions, leading to improved performance in downstream tasks such as land cover classification. 
\end{abstract}

%%%%%%%%% BODY TEXT
\input{introduction}

\input{related_work}
\input{problem_definition}
\input{dataset}

\input{methods}

\input{results}

\input{conclusion}

{\small
\bibliographystyle{plain}
\bibliography{egbib}
}

\end{document}

%% file: introduction.tex
\section{Introduction}
\begin{figure*}[!h]
    \centering
    \includegraphics[width=1.\linewidth]{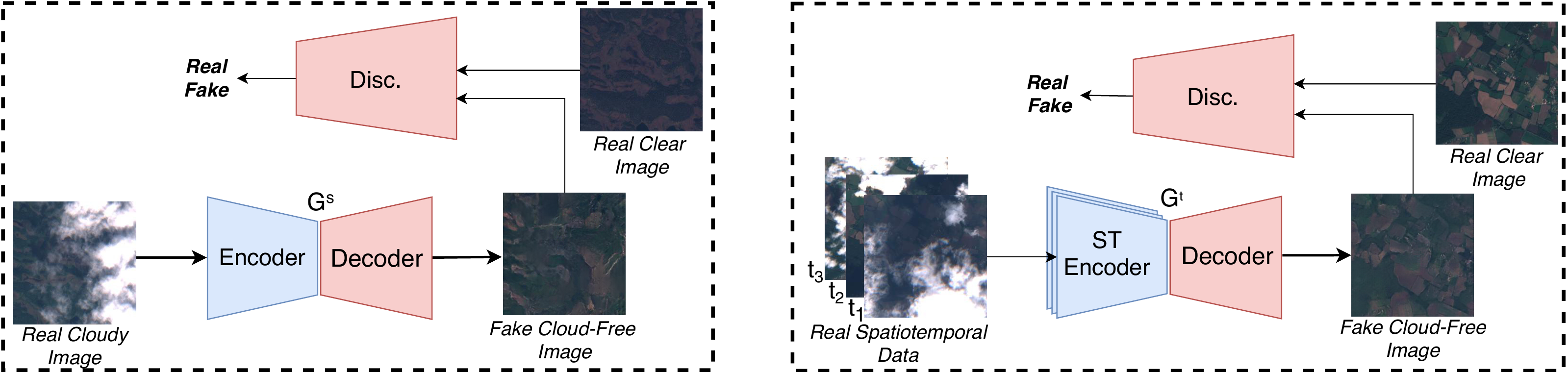}
    \caption{Full Cloud Removal Pipeline. \textbf{Left}: The first model uses our novel, real, paired dataset to generate cloud-free images (one-to-one mapping). \textbf{Right}: The final proposed model uses a novel spatiotemporal generator network to generate cloud-free images from the given sequence of cloudy images (many-to-one mapping).}
    \label{FullPipeline}
\end{figure*}

%\vishnu{This paragraph introduces why removing cloud occlusions is important}
Satellite images are increasingly utilized in a variety of applications,
including monitoring environments \cite{vakalopoulou2015building,uzkent2018tracking,uzkent2017aerial,uzkent2016real}, mapping economic development \cite{jean2016combining,sheehan2019predicting} and crop types \cite{kussul2015regional,uzkent2019learning}, classifying land cover \cite{kussul2017deep}, and measuring leaf index \cite{castillo2017estimation}. However, satellite images are often occluded by clouds -- roughly \emph{two thirds} of the world is covered by clouds at any point in time~\cite{king2013spatial}. 
Thick clouds can hide the contents of an image, and even thin translucent clouds can dramatically impact the utility of satellite images by distorting the ground below. 
Thus, removing cloud occlusions from satellite images to generate cloud-free images is a critical first step in most satellite analytic pipelines (see Fig.~\ref{FullPipeline}). 

%\vishnu{This paragraph discusses drawbacks and limitations of classical methods to motivate our solution}
Traditional approaches to removing cloud occlusions employ hand-crafted filters such as mean and median filters to generate a background image using large volume of images over a specific area \cite{helmer2005cloud, tseng2008automatic, ramoino2017ten}. For instance, \cite{ramoino2017ten} uses Sentinel-2 images taken every 6-7 days across a time period of three months. However, these image composite approaches require a large volume of mostly cloud-free images taken over a unchanging landscape, greatly limiting their usability and applications. They are also untenable in situations where the landscape gradually changes over a long period of time, as the older images do not accurately reflect the current landscape. These methods also fail to synthesize realistic ground in situations where regions are persistently occluded. Another approach leverages existing cloud-detection algorithms \cite{infrared, le2009use, hollstein2016ready} to detect cloudy regions and fill in or reconstruct those areas. However, this in-painting method fails to utilize sources of partial information--areas in shadow are partly visible, and the clouds themselves have some degree of transparency.

%\vishnu{Pluses but also drawbacks of previous generative models} 
Compared to previous methods, generative modeling has proven to be a more effective method for recovering missing information based on a learned distribution. Generative models have recently achieved state-of-the-art results in the task of image-to-image translation \cite{goodfellow2014generative, pix2pix, zhu2017toward, zhang2017image, CycleGAN, qian2018attentive, li2019single} and can be effectively applied to translate cloudy images to cloud-free images. They can be trained with many fewer images than image composite methods (ex. \cite{helmer2005cloud,tseng2008automatic,ramoino2017ten}), and the machine learning approach leverages learned as opposed to hand-crafted featurizers. However, image-to-image translation requires a paired dataset where each entry is a cloudy image and its cloud-free counterpart. Due to a lack of suitable cloudy-cloudy free paired datasets, previous attempts at applying generative models to the removal of cloud occlusions, such as in \cite{enomoto2017filmy} and \cite{singh2018cloud}, have relied on synthetically-generated image pairs where simulated clouds are artificially added to cloud-free images. Synthetic images tend to lend themselves to unrealistic representations as the simulated clouds are algorithm-based, and models trained on synthetically generated pairs fail to generalize to real images \cite{enomoto2017filmy}. Furthermore, current work using generative models does not utilize the spatiotemporal information offered by satellite imagery, relying on just a single cloudy image instead of multiple cloudy images with different cloud coverages. As a result, generated images often lack detail and specificity in previously-occluded regions and are unsuitable for downstream use.

%\vishnu{Overview of construction of dataset}
To overcome the limitations of synthetic data and the inability of previous works to utilize temporal data, we curate and assemble two new paired datasets from publicly-available Sentinel-2 satellite images \cite{helber2017eurosat}. The Sentinel-2 satellites visits the same locations periodically, with an average revisit time of approximately 6-7 days. This often allows us to find a cloud-free image in a given location as well as corresponding cloudy images from previous proximal satellite passes. Using this procedure, we construct two datasets. The first dataset contains nearly 100,000 paired images and is by far the largest paired cloud-removal dataset available for public use. The second dataset is a temporal dataset, pairing clear images with several corresponding cloudy images from different points in time to leverage the temporal nature of satellite data. Both datasets offer both RGB and infrared (IR) channels, and the temporal dataset is, to the best of our knowledge, the first of its kind. 

%\vishnu{Creation of new model}
The added information from the temporal dataset allows us to learn models that generate more detailed and accurate images, particularly in occluded areas. In order to utilize the gathered spatiotemporal information and better approximate the true cloud-free image, we design and propose several novel branched generative architectures based on U-Net \cite{ronneberger2015u} and ResNet \cite{he2016deep}. These architectures handle temporal information by extracting features from multiple images at the same location at once, then use all of the extracted features to generate a single cloud-free input. Using these new spatiotemporal generator networks (STGAN), we are able to effectively synthesize partial information from several sources into a single, detail-rich cloud-free image. 

Therefore, to overcome the limitations of previous approaches, we propose two key contributions:
\begin{enumerate}
    \item Two new paired datasets (single-image and spatiotemporal) using real-world Sentinel-2 satellite images. With both RGB and IR data, these datasets are the largest available to date.
    \item Novel spatiotemporal generator networks (STGAN) (Fig.~\ref{FullPipeline}) to better capture correlations across multiple images over an area. These models leverage our unique temporal dataset and multi-channel information offered by satellite images to effectively generate a cloud-free image. 
\end{enumerate}

%% file: related_work.tex
\section{Related Work} \label{related_work}

\textbf{Generative models for domain translation}
Deep generative models have been extensively applied to the domain translation problem. There are two approaches: unpaired image translation where there are two general categories of images and paired image translation where each image in a category directly corresponds to another image in the other category. Prominently, CycleGAN \cite{CycleGAN}, which consists of two generator-discriminator pairs: one for each image translation direction, has been used for \textit{unpaired} domain-translation applications \cite{singh2018cloud}. When using \textit{paired} images, the Pix2Pix model,  \cite{pix2pix}, a conditional-GAN (cGAN) with a U-Net \cite{ronneberger2015u} generator, has been shown to achieve state-of-the-art results across a variety of domains \cite{pix2pix}.

\textbf{Conditional generative models for cloud removal}
%\vishnu{unpaired approaches}
There has been limited work on using generative models to remove cloud occlusions. Singh et. al. \cite{singh2018cloud} attempt at \emph{unpaired image translation} using a variation on the CycleGAN~\cite{CycleGAN} to target the removal of extremely thin, filmy clouds. However, this study lacks quantitative evaluation, is limited to a very narrow scope of clouds, and is unable to tackle thicker, more opaque occlusions.

%\vishnu{paired synthetic approaches}
Approaches relying on paired image translations have largely relied on creating synthetic image pairs. Enomoto et. al. \cite{enomoto2017filmy} use Perlin noise \cite{perlin} to generate synthetic cloudy images. Using both the synthetic data and the original images' near infrared (NIR) channel, they train a Multi-spectral conditional Generative Adversarial Network (MCGAN) to generate cloud-free images. However, when creating the synthetic images, they do not modify the NIR images resulting in a mismatch in channel data \cite{infrared}. Consequentially, there is a significant difference between the MCGAN's performance on the training images in \cite{enomoto2017filmy} and the performance on real-world cloudy images, severely limiting generalizablity. Sandhan et. al. \cite{sandhan2017simultaneous} specifically target the removal of extremely filmy high-altitude clouds, using a generative model trained on synthetic data. While successful on thin and translucent clouds, their model fails to generalize to more heavily occluded images. Shibata et al. \cite{shibata2018restoration} train a model to remove cloud occlusions from sea temperature satellite imagery using pairs where the "cloud-free" image is partially covered in clouds, and a second synthetically-generated "cloudy" image takes the first image and adds additional occlusions. In this case, the GAN's objective is to only remove the additional synthetic clouds. However, this methodology is only used for reconstructing low-resolution single-channel temperature data, is drastically different from the color signals in traditional satellite imagery, and it fails to leverage partial information available in filmy regions of clouds. Here we note that there does exists a public paired dataset for cloud occlusion removal created by Lin et al. \cite{clouddataset}, but it consists of only 500 paired cloudy and cloud-free images that are geographically and topographically homogeneous.

%\vishnu{General critique of synthetic approaches}
Overall, the approaches have two main issues: they fail to generalize to real-world images and they disregard valuable temporal information available in satellite image data.

%% file: problem_definition.tex
\section{Problem Definition}
\label{sect:problem_definition}

Let $\mathcal{X} = \mathbb{R}^{w \times h \times C}$ denote the set of multispectral satellite images of size $(w,h)=256 \times 256$ with $C=4$ channels (bands).

Let $\{X_\ell^t,Z_\ell^t\}_{t,\ell}$ denote a collection of random variables $X_\ell^t, Z_\ell^t \in \mathcal{X}$, where $(X_\ell^t,Z_\ell^t)$ represent a pair of clear and cloudy views of location $\ell$ at time $t=0, 1, \cdots$.
There is a joint underlying probability distribution $p(\{X_\ell^t,Z_\ell^t\}_{t,\ell})$ describing on-the-ground changes over time and the relationship between cloud-free and cloudy satellite images. We make the following assumptions:
%\[
%p(\{X_\ell^t,Z_\ell^t\}_{t,\ell}) = \prod_t p(\{X_\ell\}^t | \{X_\ell\}^{t-1}) \prod_\ell p(Z_\ell^t |X_\ell^t)
%\]
\begin{itemize}
\item We assume $X_\ell^t$ changes slowly over time, i.e., $X_\ell^t$ is close to $X_\ell^{t-1}$ for all $t$ and for all locations $\ell$.
\item We assume $P(X_\ell^t | Z_\ell^t) = P(X_\ell^t | Z_\ell^t) \forall t, \forall \ell$, i.e., the effect of cloud cover is the same over time and at different locations.
%(probably also for all $\ell$?).
\end{itemize}

Our goal is to learn a model of the conditional distribution $P(X_\ell^t | Z_\ell^t, \cdots,Z_\ell^{t-T})$. In particular, when given a single cloudy image ($T=0$), we model the conditional distribution $P(X_\ell^t | Z_\ell^t)$. 

The major challenge associated with removing cloud occlusions is that, for every location, at time $\ell,t$ we only get to observe $X_\ell^t$ or $Z_\ell^t$, but never both.  This makes learning $P(X_\ell^t | Z_\ell^t)$ difficult.

%% file: dataset.tex
\section{Building Datasets}
To enable models to generate cloud-free images, we introduce two novel datasets. The first, $\mathcal{Y}_{single}=(X_\ell^t, Z_\ell^{t-1})$, contains single-image cloudy-cloud free pairs, and can be used to learn models in domains where only a single cloudy image is available. The second, a temporal dataset, $\mathcal{Y}_{temporal}=(X_\ell^t, Z_\ell^{t-1}, \cdots,Z_\ell^{t-T})$, contains $3$ cloudy images corresponding to each cloud-free image. 
%Both datasets contain \emph{RGB} channels and \emph{infrared} (IR) channel, offering greater flexibility than an approach dependent on more spectral channels. \vishnu{unnecessary now that we defined there being 4 channels in problem statement}

\begin{figure}[!h]
\centering
\includegraphics[width=0.45\textwidth]{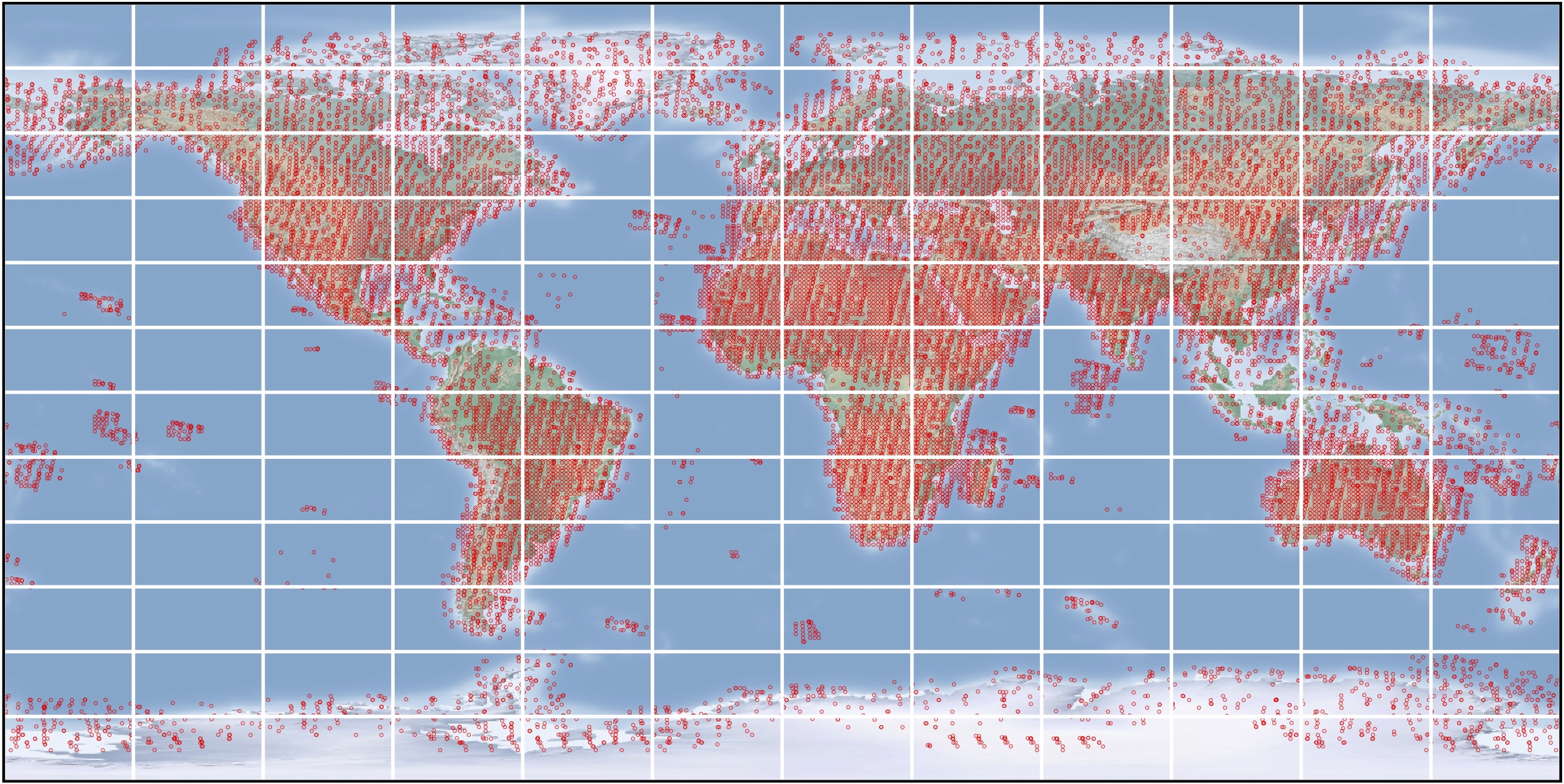}
\label{fig:lon}
\caption{Distributions of the latitude and longitude of Sentinel-2 images in our datasets.}
\label{fig:latlon}
\end{figure}

\subsection{Data Collection}
\label{DownloadData}

We use publicly-available Sentinel-2 images from 32270 distinct \emph{tiles}, where each tile is a 10980$\times$10980-pixel image with a resolution of 10m/pixel. The captured images offer multi-spectral image data from across 13 different bands, with a new image captured over the same location every $6$ days on average. We use only data from the RGB and IR bands and train all models both including and excluding the IR channel to allow use in settings where IR data is unavailable. 

\subsection{Image Crop Classification}
\label{makeReal}
To generate our dataset, we extract 100 256$\times$256 pixels crops from each of the 10980$\times$10980-pixel tiles for a total of 3,227,000 possible image locations in the dataset. We then label each image crop as \emph{clear} or \emph{cloudy} by first thresholding based on percentage of cloud cover, which we obtain from the cloud collection algorithm presented in \cite{hollstein2016ready}, and then applying heuristics as described in the appendix. We retain image crops with percentage cloud cover under 1\% as clear and crops with percentage cloud cover between 10\% and 30\% as cloudy. We exclude images with higher than 30\% cloud cover, as many of the images had insufficient visible ground upon manual inspection. We restrict images of ocean to no more than 10\% of the dataset, discarding additional images. The distribution of the locations of image crops in the resulting dataset is displayed in Fig.~\ref{fig:latlon}. Further details on image classification are included in the supplemental appendix.

\subsection{Image Crop Pairing}
\label{makePairs}
In order to find a clear image at every location, we find the most recent clear crop at a given location, $X_\ell^t$, using the clear/cloudy image crop labels from Sect.~\ref{makeReal}. For each clear crop, $X_\ell^t$, we find all associated cloudy crops, potentially $Z_\ell^{t-1}, \cdots,Z_\ell^{t-2}, \cdots$, taken from the same location in the prior $35$ days.
We build $\mathcal{Y}_{single}$ with only the most recent cloudy image as input. In $\mathcal{Y}_{single}$, there are 97640 image pairs, drawn from 17800 distinct tiles worldwide. Each pair is of the form $(X_\ell^t,Z_\ell^{t-1})$. Under our assumptions, $X_\ell^{t-1}$ is an accurate approximation of $X_\ell^t$, so the pair $(X_\ell^t,Z_\ell^{t-1})$ approximates the pair $(X_\ell^{t-1},Z_\ell^{t-1})$. This dataset helps us model the conditional distribution $P(X_\ell^t | Z_\ell^t)$. 

Next, we build $\mathcal{Y}_{temporal}$ using clear images that are paired with several cloudy images. In $\mathcal{Y}_{temporal}$, we pair each clear image with the $3$ most recent cloudy images. Discarding clear images that do not have at least three corresponding cloudy images, there are 3101 images, drawn from 945 distinct tiles worldwide. Again, the grouped images are of the form $(X_\ell^t, Z_\ell^{t-1}, \cdots,Z_\ell^{t-T})$ and help approximate $P(X_\ell^{t-1} | Z_\ell^{t-1}, \cdots,Z_\ell^{t-T})$. 

\begin{figure}[!h]
    \centering
    \includegraphics[width=0.4\textwidth]{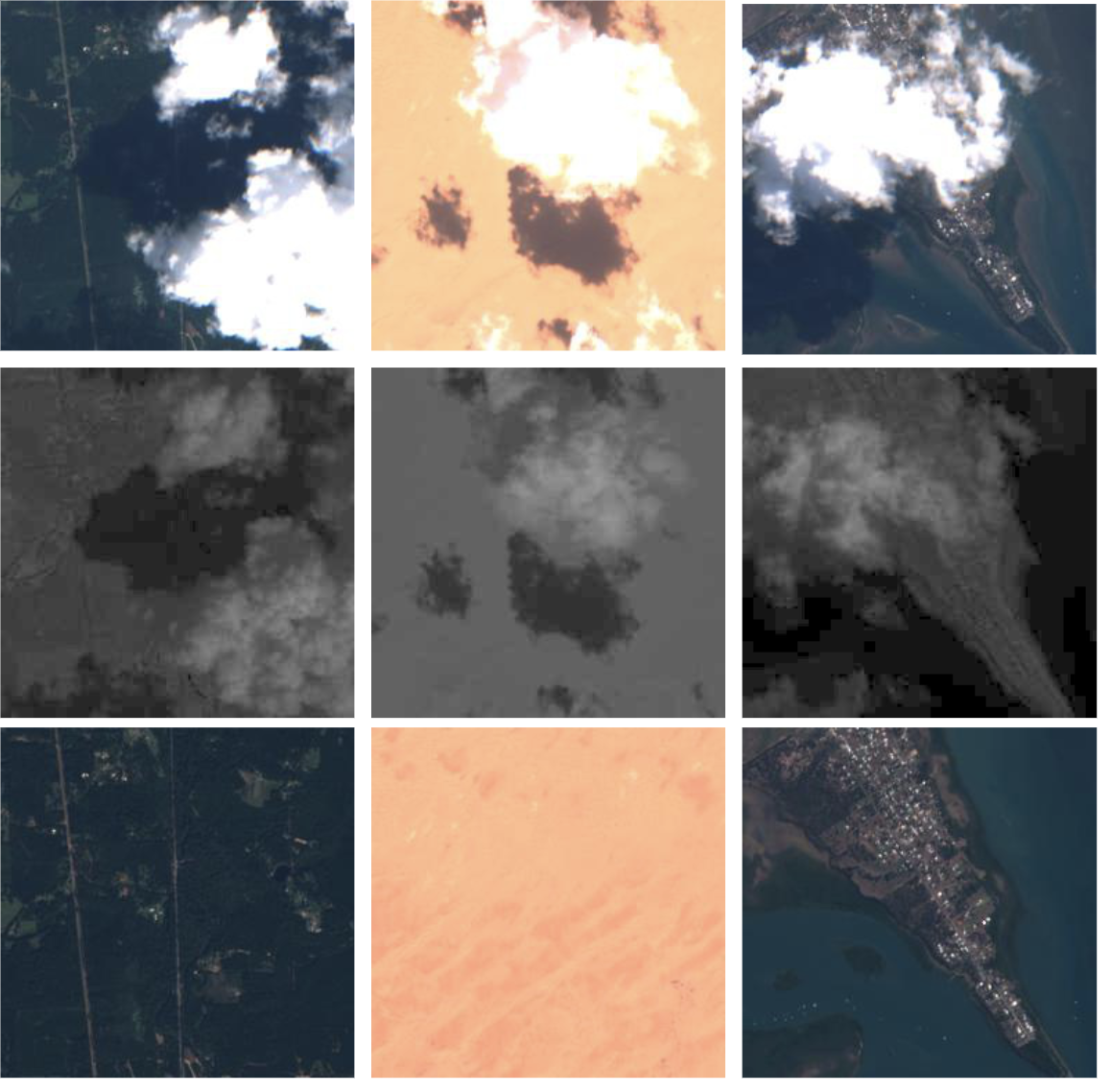}
    \caption{Samples from our single image paired dataset. The first row displays the RGB channels of the cloudy image, $Z_\ell^t$, the second row displays the IR channel, and the third row displays clear RGB images, $X_\ell^t$.}
    \label{fig:single_paired_dataset}
\end{figure}

%% file: methods.tex
\section{Methods} 
\subsection{Cloud Removal using a Single Image}
\label{sect:pix2pix}
We approximate $P(X_\ell^t | Z_\ell^t)$ using a Pix2Pix model~\cite{pix2pix}, which has achieved state-of-the-art results for paired image to image translation tasks. Pix2Pix is an instance of a conditional GAN (cGAN) where the generator has a variant on an encoder-decoder structure, with a series of convolutional layers compressing, then expanding the input image. The Pix2Pix generator is unusual since the generator has a number of skip connections, as in the U-Net \cite{ronneberger2015u} architecture. This allows the network to bypass further encoder/decoder layers if the model determines that additional compression and decompression is not necessary. Further details on our implementation of the Pix2Pix model are available in the supplemental appendix.

We train the Pix2Pix model on the paired dataset $\mathcal{Y}_{single}$, both including and excluding IR information, and evaluate on real-world cloudy and cloud-free images. The model trained with IR information can be directly compared to the MCGAN approached discussed earlier \cite{enomoto2017filmy}, as both models use RGB and IR channels as input.

\subsection{Cloud Removal using Spatiotemporal Information} \label{stgan}
\begin{figure*}[!h]
\centering
    \subfloat[]{\includegraphics[width=0.48\textwidth]{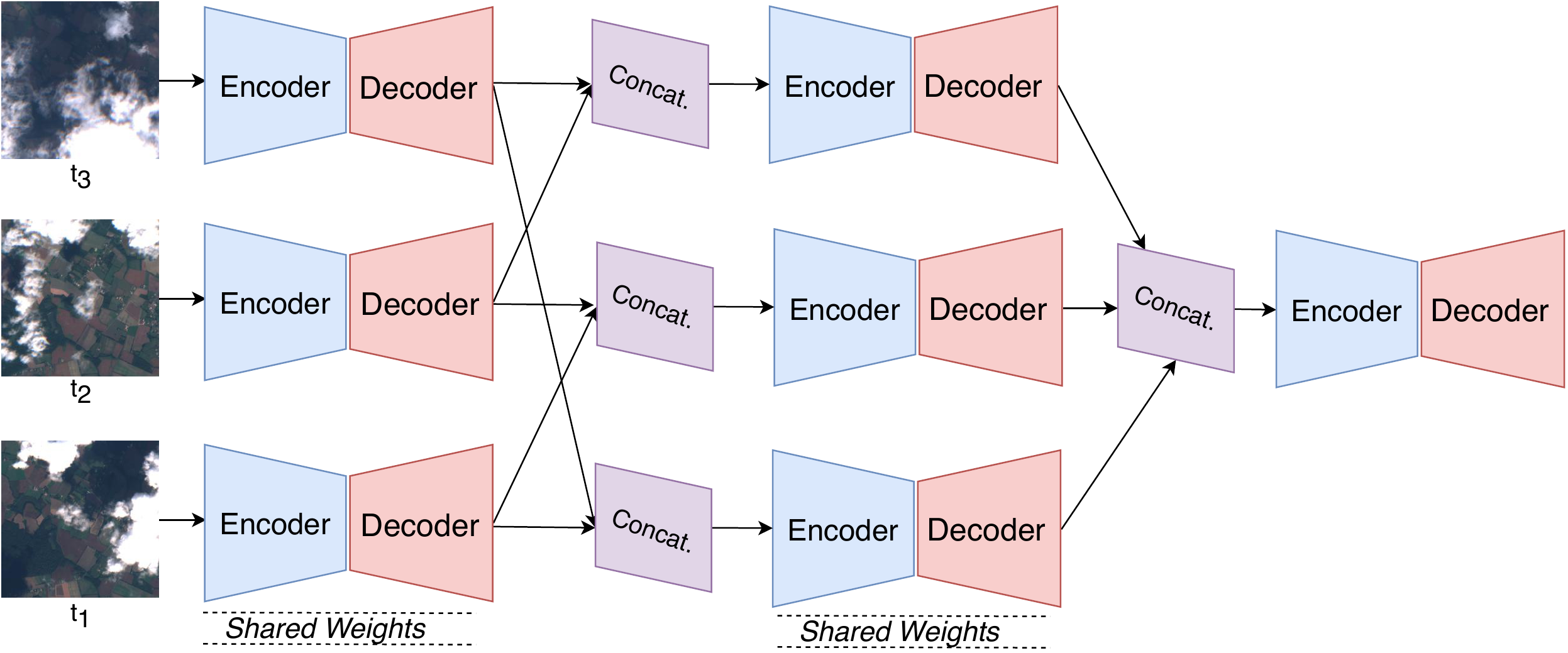}
    \label{fig:spatiotemporal_network_II}}
    \quad
    \subfloat[]{\includegraphics[width=0.48\textwidth]{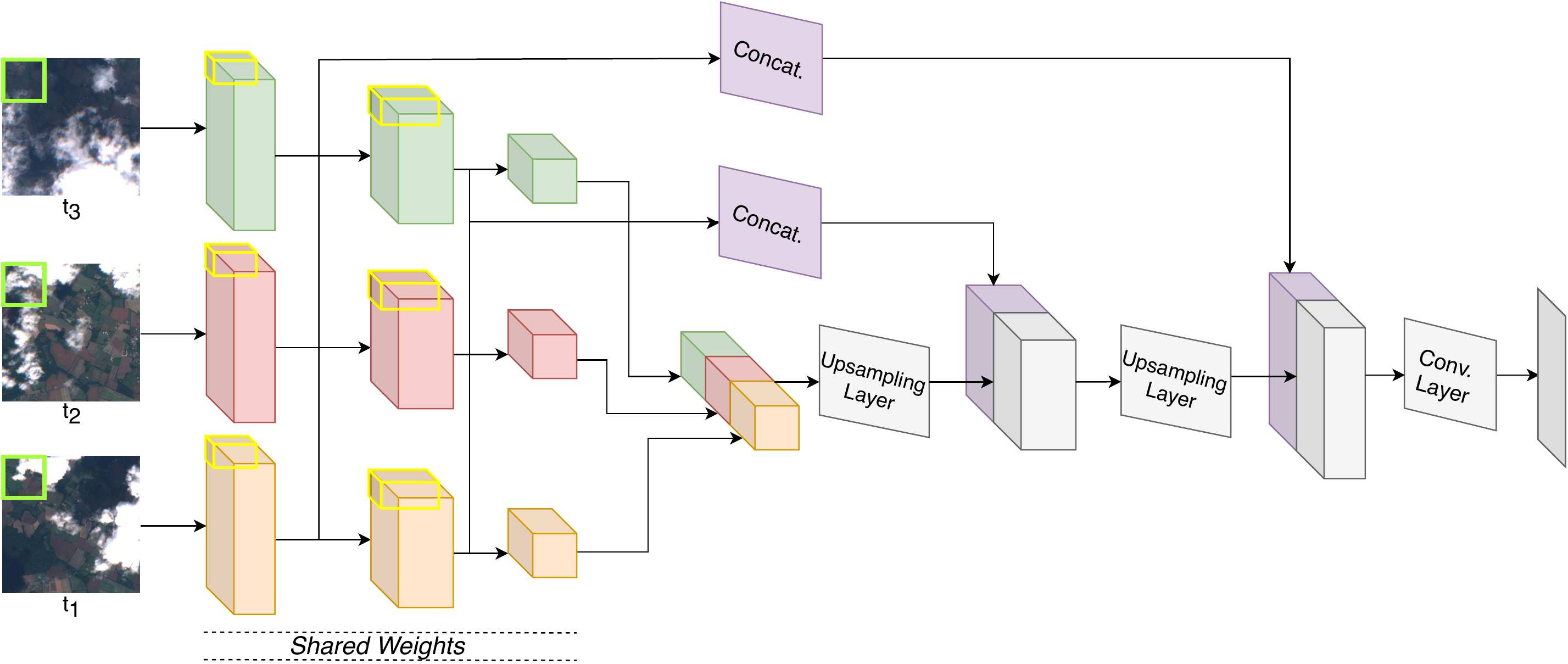}
    \label{fig:spatiotemporal_network_III}
    }
    \caption{The proposed \emph{spatiotemporal} generator networks. \textbf{Left:} The Branched ResNet generator uses encoder-decoder residual networks to learn feature maps from individual images, then repeatedly combines these feature maps and passes them through subsequent encoder-decoders to extract features from multiple images. \textbf{Right:} The Branched U-Net modifies a traditional U-Net architecture to allow each of the input images to encode separately, and then they are decoded together. }
    \label{fig:spatiotemporal_networks}
\end{figure*}

We train our spatiotemporal model on $\mathcal{Y}_{temporal}$, described in \ref{makePairs} to learn the distribution $P(X_\ell^t | Z_\ell^t, \cdots,Z_\ell^{t-T})$. As with the single-image model, we train the spatiotemporal model on versions of the dataset both including and excluding an IR channel.

To fully utilize and incorporate the spatiotemporal information, we propose two novel generator architectures, paired with the PatchGan discriminator from \cite{pix2pix}:
\begin{enumerate}
    \item A \emph{branched} ResNet model that first passes the three inputs through separate encoder-decoder pipelines, then concatenates the three sets of image features in a pairwise manner to produce three new inputs, which in turn are each passed through separate encoder-decoder pipelines and are then concatenated, before being passed through one final encoder-decoder pipeline. This model is shown in Fig.~\ref{fig:spatiotemporal_network_II}. 
    \item A U-Net where each image is passed into a separate encoder pipeline, then concatenated and fed into a single composite decoder, as shown in Fig.~\ref{fig:spatiotemporal_network_III}. 
\end{enumerate}

\textbf{Branched ResNet} For this architecture, we define each of the individual encoder-decoder pipelines as follows: two convolutional layers with stride $2$ to downsample the feature map, followed by nine residual convolutional layers, followed by two convolutional layers with stride $1/2$ to upsample the feature map. The inspiration for the architecture of the encoder-decoder comes from \cite{johnson2016perceptual}, and was previously shown to perform well in CycleGAN~\cite{CycleGAN}. The intuition between using multiple encoder-decoder pipelines is that the first stage of pipelines extracts the network's best \emph{guess} of the true cloud-free image from individual cloudy images, then the second stage of pipelines combines the guesses (i.e. convolutional volumes) pairwise for each of the branches, and the final encoder-decoder pipeline combines all the guesses into one final output generation. The network maximizes information gained from using multiple images by providing pairwise overlap between the convolutional volumes in a branch-like manner.

\textbf{Branched U-Net} For this architecture, similar to \cite{pix2pix}, we build upon the general U-Net framework with 8 downsampling and 8 upsampling blocks, with skip connections added between each block $i$ and block $n - i$, where $n = 16$ is the total number of blocks. Each block, as in \cite{pix2pix}, is comprised of a convolutional layer (downsampling or upsampling depending on the stride), followed by batch normalization and a ReLU unit. Three input images are encoded in separate pipelines, where each block is connected to the corresponding block in the decoder by a skip connection. After the final encoding block, the three feature maps are concatenated and passed to a single decoder pipeline. After each decoder block, the output is concatenated with outputs from each of three encoder blocks through skip connections from each of the encoder pipelines. The intuition behind this architecture involves encoding images separately to extract key features from each of them, while decoding the images together in order to generate a single cohesive output image. The model architectures are further described in the supplementary appendix.

The objective function of the spatiotemporal models consists of a conditional GAN loss, and L1 loss which is parameterized by the hyperparameter $\lambda$:
\begin{align*}
    \mathcal{L}_{cGAN}(G, D) &= \mathbb{E}_{(X_\ell^t,Z_\ell)} [\log D(Z_\ell,X_\ell^t)] \\
    &+ \mathbb{E}_{x_{t}^{c}}[log(1-D(Z_\ell),G(x_{t}^{c}))] \\ 
    \mathcal{L}_{L1}(G) &= \mathbb{E}_{(Z_\ell,X_\ell^t)} [||X_\ell^t-G(Z_\ell)||_{1}] \\ 
    G^{{*}} &= arg \min_{D^{s}} \max_{G} \mathcal{L}_{cGAN}(G, D) + \lambda \mathcal{L}_{L1}(G)
\end{align*}
where $G$, and $D$ represent the spatiotemporal generator and discriminator networks, and $Z_\ell=Z_\ell^{t-1}, \cdots,Z_\ell^{t-T}$. The input to the discriminator, $D$, is a clear image, $X_\ell^t$, or a fake clear image, $\hat{X_\ell^t}$, generated by $G$ passed along with the original cloudy images, $Z_\ell^t, \cdots,Z_\ell^{t-T}$. 

%% file: results.tex
\section{Experiments and Results}

We train both Pix2Pix and MCGAN models on our single image dataset ($\mathcal{Y}_{single}$) and train our novel STGAN architectures on our spatiotemporal dataset ($\mathcal{Y}_{temporal}$). In order to effectively evaluate the model performance on removal of cloud occlusions, we compare against traditional approaches (mean filter, median filter, composite filter) and report image similarity metrics and downstream task performance. Ultimately, we are able to achieve state-of-the-art results for cloud occlusion removal using both of our newly constructed datasets and STGAN architectures.

\subsection{Metrics}
Two standard metrics used in measuring image similarity and degradation are peak signal-to-noise ratio (PSNR) and structural similarity index (SSIM) \cite{wang2004image}. 
PSNR, largely based on mean-squared error (MSE), is a metric that is based on the average difference between corresponding pixels in two images. It is defined as follows:

\begin{equation}
    \mathit{PSNR(x,y)} = 10 \cdot \log_{10} \left( \frac{\mathit{MAX}_I^2}{\mathit{MSE(x,y)}} \right)
\end{equation}

Here, $x$ and $y$ are two images, $m$ and $n$ are the height and width of the images respectively, $MSE$ is the mean-squared error between the two images, and $\mathit{MAX}_I^2$ is the maximum possible value of any pixel in the image. PSNR values range from $0$ to $48$ in images where the maximum pixel value is $255$, with larger values representing more similar images. SSIM is a metric that, unlike PSNR, aims to track similarity in visible structures in the image and captures more of the relationship between large-scale features of the image than PSNR, which is pixel-based. SSIM values range from $0$ to $1$, with larger values representing more similar images.  
 
\subsection{Experimental Details} \label{exp_details}

We train the Pix2Pix model \cite{pix2pix} on $\mathcal{Y}_{single}$ with both RB and RGB+IR data. Additionally, we use the previous state-of-the-art model for removing all types of cloud occlusions, MCGAN, as a baseline. We train the MCGAN \cite{enomoto2017filmy} model from scratch on our paired dataset ($\mathcal{Y}_{single}$) and use it as a point of comparison for our own proposed models.

Similarly, we train the proposed STGAN models on $\mathcal{Y}_{temporal}$ for both RGB and RGB+IR data. For the branched models from Fig.~\ref{fig:spatiotemporal_networks}, we train models both sharing weights across different branches and with independent weights across branches. We compare against two baselines which aggregate the three cloudy images in varying ways: a mean filter where the generated image is the mean of the three temporal cloudy images and a median filter where the generated image is the median of all cloudy images. 

The dataset splits, training details and hyperparameters can be referenced in the supplementary appendix. 
%For each of the models, we tune the L1-element of the generator loss and the batch size of the input through a random grid search.  We train our models for 200 epochs, using the Adam optimizer \cite{kingma2014adam} with a momentum of 0.5 and a beta of 0.99. The learning rate of 1e-3 was kept the same for the first 100 epochs and then linearly decayed to zero over the next 100. We determine the optimal hyper-parameters following a random grid search and used the models with highest SSIM in validation.%
%\ej{Is any of this giving anyone useful information? It feels like you can dump most of it in the appendices, but I'm not sure what's standard for vision papers. Also Experimental Details is better than procedures, because this isn't a procedure.} \vishnu{I think we should keep this? But not 100\% sure.}%

\subsection{Results on Single Image Models}
\begin{figure}[!h]
    \centering
    \includegraphics[width=0.5\textwidth]{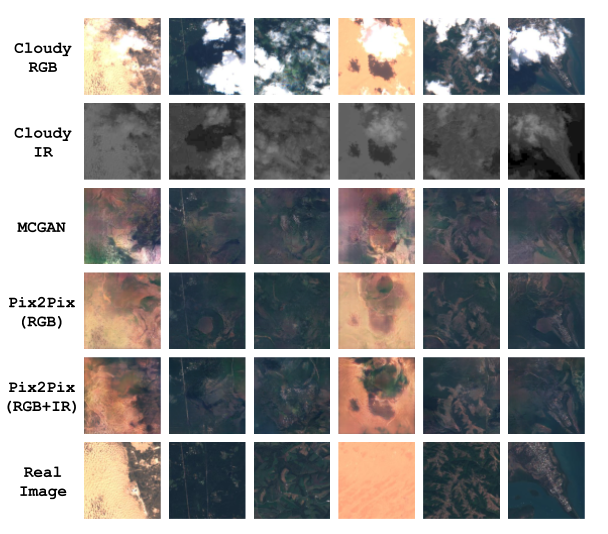}
    \caption{Results of models trained on $Y_{single}$ to generate cloud-free images given an input cloudy image. Dense clouds make it challenging to learn mapping by using a single image.}
    \label{fig:single_image_results}
\end{figure}

\textbf{Qualitative Results} Fig.~\ref{fig:single_image_results} shows the original Sentinel-2 cloudy images, the ground-truth cloud-free images, images generated by the MCGAN baseline \cite{enomoto2017filmy}, and cloud-free images generated using the Pix2Pix model. We can see that the Pix2Pix model is able to generate the parts of the image where clouds and their shadows are not present and keep those areas intact while making reasonable, if sometimes blurry, inferences about the ground beneath the occlusions. On the other hand, the MCGAN, behaves unrealistically in some cases, even failing to preserve some visible areas. 

\textbf{Quantitative  Results} Table \ref{table:single_image_models} shows that the Pix2Pix model trained on the single image dataset ($\mathcal{Y}_{single}$) achieves state of the art results when trained on either RGB data or RGB+IR data. In particular, note that both models outperform the baseline and previous state-of-the-art MCGAN: the Pix2Pix RGB-trained model's PSNR of 21.130 outperformed MCGAN's 20.871, and the Pix2Pix RGB+IR-trained model's SSIM of 0.485 vastly outperformed MCGAN's 0.424. Interestingly, the IR data leads to a greater change in SSIM than PSNR, which indicates better preservation of visual structures (see Fig.~\ref{fig:effect_of_ir}). In summary, our single-image model outperforms existing work and, to the best of our knowledge, achieves state of the art results at removing cloud occlusions from single-image inputs. 

\begin{table}[]
\centering
\begin{tabular}{@{}lllll@{}}
\toprule
& \multicolumn{2}{c}{Validation Set} & \multicolumn{2}{c}{Test Set} \\ \midrule
Models & PSNR & SSIM & PSNR & SSIM \\
\begin{tabular}[c]{@{}l@{}}Pix2Pix (RGB) \end{tabular} & \textbf{23.130} & 0.442 & \textbf{22.894} & 0.437 \\
\begin{tabular}[c]{@{}l@{}}Pix2Pix (RGB + IR)\end{tabular} & 21.352 & \textbf{0.485} & 21.146 & \textbf{0.481} \\ 
\begin{tabular}[c]{@{}l@{}}MCGAN (RGB + IR)\end{tabular} & 20.871 & 0.424 & 21.013 & 0.381 \\ 
\begin{tabular}[c]{@{}l@{}}Raw Cloudy Images\end{tabular} & 8.742 & 0.396 & 8.778 & 0.398 \\ \bottomrule
\end{tabular}
\caption{The performance of the models, in terms of PSNR and SSIM scores, on the paired dataset with and without IR data relative to the cloud-free ground truth.}
\label{table:single_image_models}
\end{table}

\begin{table}[]
\centering
\begin{tabular}{@{}lllll@{}}
\toprule
& \multicolumn{2}{c}{Validation Set} & \multicolumn{2}{c}{Test Set} \\ \midrule
Models & PSNR & SSIM & PSNR & SSIM \\
\begin{tabular}[c]{@{}l@{}}Pix2Pix (RGB)\end{tabular} & 23.130 & 0.442 & 22.894 & 0.437 \\
\begin{tabular}[c]{@{}l@{}}Mean Filter \end{tabular} & 16.962 & 0.174 & 16.893 & 0.173  \\
\begin{tabular}[c]{@{}l@{}}Median Filter\end{tabular} & 9.081 & 0.357 & 9.674 & 0.395  \\
\begin{tabular}[c]{@{}l@{}}STGAN U-Net \end{tabular} & 25.484 & 0.534 & 25.822 & 0.564 \\
\begin{tabular}[c]{@{}l@{}}STGAN ResNet \end{tabular} & \textbf{25.519} & \textbf{0.550} & \textbf{26.000} & \textbf{0.573} \\
\begin{tabular}[c]{@{}l@{}}STGAN U-Net (IR) \end{tabular} & 25.142 & 0.651 & 25.388 & 0.661 \\
\begin{tabular}[c]{@{}l@{}}STGAN ResNet (IR) \end{tabular} & \textbf{25.628} & \textbf{0.724} & \textbf{26.186} & \textbf{0.734} \\
\begin{tabular}[c]{@{}l@{}}Raw Cloudy Images\end{tabular} & 7.926 & 0.389 & 8.289 & 0.422 \\ 
\bottomrule
\end{tabular}
\caption{The performance of the models, in terms of PSNR and SSIM scores, on the real spatiotemporal dataset relative to the cloud-free ground truth. }
%\burak{Any reason why Pix2Pix does not use IR?}} \vishnu{It does but we included that in an earlier table and it's not needed as a point of reference for this
\label{table:spatiotemporal_models}
\end{table}

\subsection{Results on Spatiotemporal Models}

\begin{figure*}[!h]
    \centering
    \includegraphics[width=0.85\textwidth]{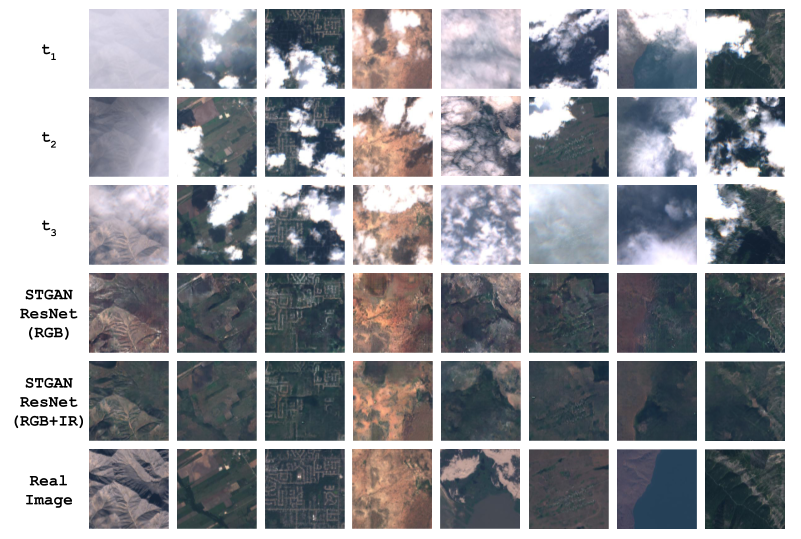}
    \caption{Results of learned many-to-one mapping to generate cloud-free images given a sequence of cloudy images. Note that a number of these cases, especially the fifth column, involve dense cloud coverage and are quite difficult to reconstruct. IR information is available for all cloudy images but is not shown.}
    \label{fig:results_examples}
\end{figure*}

\textbf{Qualitative Results}
Fig.~\ref{fig:results_examples} shows the three input cloudy images, ground-truth cloud-free image and our generated cloud-free images from our best performance model based on SSIM. Both are STGAN models based off a branched ResNet architecture. If a detail is clearly visible in at least one image, it appears in the images generated by both models. The models are even able to generate realistic ground features in some areas obscured across all cloudy images. Compared to the single-image models, the STGAN models are able to generate much more realistic and crisp cloud-free images that infer features and terrain more accurately. The models generate key characteristic features and terrain that are also present in the ground truth cloud-free images. In addition, the inclusion of IR data influences the fine details of generated images. While images generated from the RGB model tend to synthesize a variety of false artifacts, the model incorporating IR tends to preserve visible detail while filling in unknown regions with far fewer artifacts. 

Furthermore, we assess the effectiveness of composite filter where a cloud detection algorithm \cite{hollstein2016ready} is run over all cloudy images and the image is generated by averaging over all pixels that are cloud-free (i.e. an in-painting approach). Most of the images generated by the composite filter (shown in Appendix) contain large swathes of missing pixels (i.e. where there were no cloud-free pixels in all temporal images) demonstrating that it has difficulty removing dense occlusions.  

\begin{figure}[!h]
    \centering
    \includegraphics[width=0.5\textwidth]{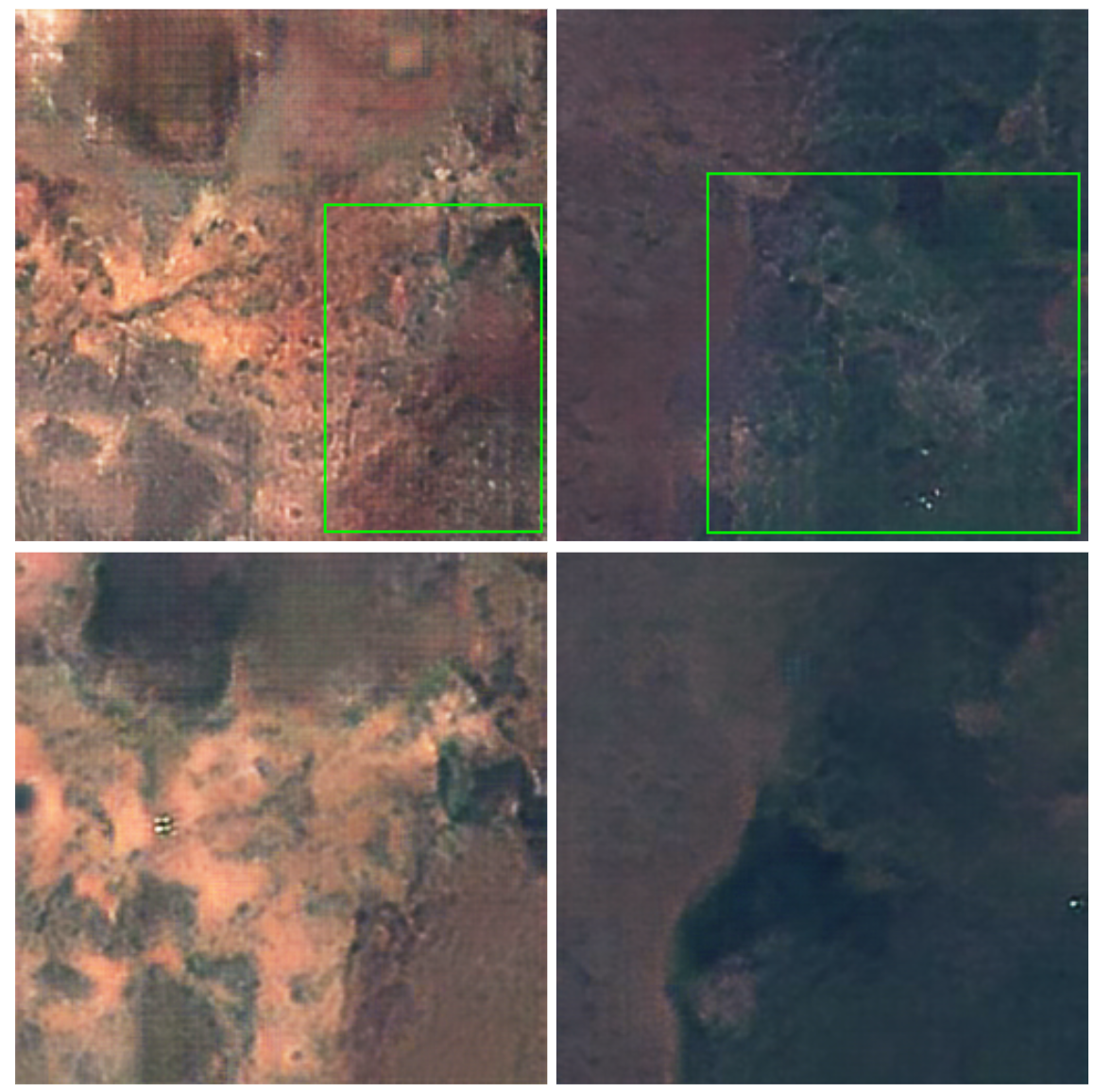}
    \caption{A closer look at the effects of including IR information (from Figure \ref{fig:results_examples}). The first row is the image generated from RGB channels, while second row includes the IR channel. Note that the models using only RGB channels tend to introduce false artifacts as highlighted above.}
    \label{fig:effect_of_ir}
\end{figure}

\textbf{Quantitative  Results}
Table~\ref{table:spatiotemporal_models} shows that the ResNet-based STGAN outperforms the U-Net-based STGAN. With RGB data, the ResNet-based STGAN achieves PSNR and SSIM values of 26.000 and 0.573 respectively, while the incorporation of IR data raises these numbers to a PSNR of 26.186 and an SSIM of 0.734, an improvement of 0.161 in SSIM over the U-Net. This is consistent with visual observations that IR data helps the model maintain clarity in unobstructed reasons. The general coloring of the image does not differ by much (see similar PSNR scores), but the details generated by the models are substantially different, as seen in Fig.~\ref{fig:effect_of_ir}.

The STGAN models far outperform rudimentary baselines, and provide a substantial improvement over the single-image Pix2Pix models. The STGAN trained on RGB data achieves a 13.6\% improvement in PSNR and a 31.1\% improvement in SSIM over the single-image model, while the STGAN which incorporates the IR band gains a 22.6\% improvement in PSNR and a 52.6\% improvement in SSIM. Improvements in SSIM are generally due to better performance in capturing the visible structures in the image. Therefore, our novel temporal architectures enable the model to capture much more detail in the output image than single-image models such as Pix2Pix. 

Note that the ResNet-based STGAN on RGB images has a PSNR of 26.373, beating both the PSNR of 17.45 from \cite{sandhan2017simultaneous} by 8.92 (despite their model only operating only on filmy cloud images) and the PSNR of 20.871 from MCGAN \cite{enomoto2017filmy} by 5.50. Thus, to the best of our knowledge, our spatiotemporal models achieve state-of-the-art results on removing dense cloud occlusions from satellite images with both RGB and RGB+IR channels.

\subsection{Evaluation on Downstream Tasks}
Finally, we evaluate the performance of models trained on our datasets on a downstream task of land cover classification. We train a baseline land classification model on the EuroSat dataset \cite{eurosat}, a pre-labeled dataset comprised of 27,000 labeled Sentinel-2 satellite images consisting of 10 classes (sea and lake, river, residential, permanent crop, pasture, industrial, highway, herbaceous vegetation, forest and annual crop). From our datasets, we hand-labeled images (corresponding cloudy and cloud-free) with approximately equal distribution across all 10 categories to create the test set and generated cloud-free images from the test set using our proposed approaches, as described in the Appendix. Finally, we evaluate the accuracy of the trained land classification model (on the test set) in predicting the correct class from 1) true cloudy images 2) true cloud-free images and 3) generated cloud-free images.  

\begin{table}[]
\centering
\begin{tabular}[c]{ll}
\toprule
Model              & Accuracy \\
\midrule
Cloudy             & 72.48\%        \\
Cloud-free         & 98.66\%        \\
MCGAN              & 88.59\%        \\
Pix2Pix (RGB)      & 90.60\%        \\
Pix2Pix (RGB + IR) & 91.27\%        \\
STGAN (RGB)        & 93.96\%        \\
STGAN (RGB + IR)   & 93.96\%        \\
\bottomrule
\end{tabular}
\caption{Effectiveness of generated cloud-free images compared to true cloud-free and cloudy images for the task of land cover classification}
\label{table:downstream_task}
\end{table}

As seen in Table \ref{table:downstream_task}, the STGAN generated cloud-free images that perform comparably (93.96\%) against the true cloud-free images (98.66\%). Note that the classification categories are explicit and include labels with subtle visual differences such as industrial, highway, and residential land areas which require detailed images to be classified correctly. Therefore, models trained using our generated images generalize to real data. In contrast, the cloudy images have a significantly worse accuracy of 72.48\% than both the true cloud-free and generated cloud-free images. Thus, in alignment with our reported metrics, the spatiotemporal models perform better than single-image models, which outperform the previous state of the art, MCGAN.

With access to cloud-free satellite images, automated land cover classification can be used towards agriculture, disaster recovery, climate change, urban development, and environmental monitoring \cite{lc_app1}, \cite{lc_app2}, \cite{lc_app3}, \cite{lc_app4}. Therefore, this downstream task evaluation demonstrates that the proposed models, trained on our datasets, can accurately reconstruct cloud-free images from cloudy images for land cover classification, which has many critical applications.

%% file: conclusion.tex
\section{Conclusion}
In this study, we propose a novel framework to generate cloud-free images from cloudy images using deep generative models. We construct novel, large-scale, global, paired spatial and spatiotemporal datasets using publicly available Sentinel-2 images. These datasets are the largest datasets of their kind available to date. Additionally, we introduce novel generative architectures (STGAN) that leverage our spatiotemporal satellite data to recover realistic cloud-free images. Our experiments demonstrate that the STGAN significantly outperforms the state-of-the-art models on generating cloud-free images across a variety of challenging terrains, even in the cases of thick and dense cloud occlusions. Finally, we have demonstrated that the generated cloud-free images are useful for real-world downstream tasks. We hope our work makes more satellite data usable for further research and applications.